\documentclass[11pt,a4paper]{article}

\usepackage[hyperref]{emnlp2018}
\usepackage{times}
\usepackage{latexsym}
\usepackage{amsfonts}
\usepackage{multirow}
\usepackage[pdftex]{graphicx}

\usepackage{pgf}
\usepackage{tikz}

\usepackage{epstopdf}
\usepackage{url}
\aclfinalcopy 

\title{Translating a Math Word Problem to an Expression Tree}

\author{Lei Wang$^1$\Thanks{The work was done when Lei Wang and Deng Cai were interns at Tencent AI Lab.}, Yan Wang$^2$, Deng Cai$^{23*}$, Dongxiang Zhang$^1$, Xiaojiang Liu$^2$ \\
        \small$^1$Center for Future Media and School of Computer Science \& Engineering, UESTC, $^2$Tencent AI Lab\\
        \small$^3$Department of Systems Engineering and Engineering Management, The Chinese University of Hong Kong, Hong Kong  \\
        {\small \tt demolei@outlook.com, \{bradenwang, kieranliu\}@tencent.com},\\
       \small \tt thisisjcykcd@gmail.com, zhangdo@uestc.edu.cn}

\date{}
\begin{document}
\maketitle 
\begin{abstract}
  Sequence-to-sequence (\textsc{Seq2Seq}) models have been successfully applied to automatic math word problem solving. Despite its simplicity, a drawback still remains: a math word problem can be correctly solved by more than one equations. This non-deterministic transduction harms the performance of maximum likelihood estimation. In this paper, by considering the uniqueness of expression tree, we propose an equation normalization method to normalize the duplicated equations. Moreover, we analyze the performance of three popular \textsc{Seq2Seq} models on the math word problem solving. We find that each model has its own specialty in solving problems, consequently an ensemble model is then proposed to combine their advantages. Experiments on dataset Math23K show that the ensemble model with equation normalization significantly outperforms the previous state-of-the-art methods.

\end{abstract}
%

\section{Introduction}

Developing computer systems to automatically solve math word problems (MWPs) has been an interest of NLP researchers since 1963 \cite{feigenbaum1963computers,bobrow1964natural}. A typical MWP is shown in Table \ref{problem_example}. Readers are asked to infer how many pens and pencils Jessica have in total, based on the textual problem description provided. Statistical machine learning-based methods \cite{kushman2014learning,amnueypornsakul2014machine,zhou2015learn,mitra2016learning,DBLP:journals/tacl/RoyR18} and semantic parsing-based methods \cite{shi2015automatically,koncel2015parsing,roy2016solving,huang2017learning} are proposed to tackle this problem, yet they still require considerable manual efforts on feature or template designing. For more literatures about  solving math word problems automatically, refer to a recent survey paper \citet{zhang2018survey}.

Recently, the Deep Neural Networks (DNNs) have opened a new direction towards automatic MWP solving. \citet{ling2017program} take multiple-choice problems as input and automatically generate rationale text and the final choice. \citet{lei2018mathdqn} then make the first attempt of applying deep reinforcement learning to arithmetic word problem solving. \citet{wang2017deep} train a deep neural solver (DNS) that needs no hand-crafted features, using the \textsc{Seq2Seq} model to automatically learn the problem-to-equation mapping.

\begin{table}\label{problem_example}
\begin{tabular}{|p{7cm}|}
\hline \textbf{Problem}: Dan has 5 pens and 3 pencils, Jessica has 4 more pens and 2 less pencils than him. How many pens and pencils does Jessica have in total? \\\hline
 \textbf{Equation}: $x = 5 + 4 +3 -2$;\quad \textbf{Solution:} 10\\\hline
\end{tabular}
\caption{A math word problem}
\end{table}

Although promising results have been reported, the model in \citep{wang2017deep} still suffers from an equation duplication problem: a MWP can be solved by multiple equations. Taking the problem in Table \ref{problem_example} as an example, it can be solved by various equations such as $x = 5 + 4 +3-2$, $x=4+(5-2)+3$ and $x = 5 -2 + 3 +4$. This duplication problem results in a non-deterministic output space, which has a negative impact on the performance of most data-driven methods. In this paper, by considering the uniqueness of expression tree, we propose an equation normalization method to solve this problem.

Given the success of different \textsc{Seq2Seq} models on machine translation (such as recurrent encoder-decoder \cite{wu2016google}, Convolutional \textsc{Seq2Seq} model \cite{gehring2017convolutional} and Transformer \cite{vaswani2017attention}), it is promising to adapt them to MWP solving. In this paper, we compare the performance of three state-of-the-art \textsc{Seq2Seq} models on MWP solving. We observe that different models are able to correctly solve different MWPs, therefore, as a matter of course, an ensemble model is proposed to achieve higher performance. Experiments on dataset Math23K show that by adopting the equation normalization and model ensemble techniques, the accuracy boosts from 60.7\% to 68.4\%.

The remaining part of this paper is organized as follows: we first introduce the \textsc{Seq2Seq} Framework in Section 2. Then the equation normalization process is presented in Section 3, following which three \textsc{Seq2Seq} models and an ensemble model are applied to MWP solving in Section 4. The experimental results are presented in Section 5. Finally we conclude this paper in Section 6.

\section{\textsc{Seq2Seq} Framework}
The process of using \textsc{Seq2Seq} model to solve MWPs can be divided into two stages \cite{wang2017deep}. In the first stage (number mapping stage), significant numbers (numbers that will be used in real calculation) in problem $P$ are mapped to a list of number tokens $\{n_1, \dots, n_m\}$ by their natural order in the problem text. Throughout this paper, we use the significant number identification (SNI) module proposed in \cite{wang2017deep} to identify whether a number is significant. In the second stage, \textsc{Seq2Seq} models can be trained by taking the problem text as the source sequence and equation templates (equations after number mapping) as the target sequence.

Taking the problem $P$ in Table \ref{problem_example} as an example, first we can obtain a number mapping $M: \{n_1=5;\quad n_2=3;\quad n_3=4;\quad n_4=2;\}$, and transform
the given equation $E_P: x = 5 + 4 +3 -2$ to an equation template $T_P: x = n_1+n_3+n_2-n_4$. During training, the objective of our \textsc{Seq2Seq} model is to maximize the conditional probability $P(T_p | P)$, which will be decomposed to token-wise probabilities. During decoding, we use beam search to approximate the most likely equation template. After that, we replace the number tokens with actual numbers and calculate the solution $S$ with a math solver.
\section{Equation Normalization}

In the number mapping stage, the equations $E_P$ have been successfully transformed to equation templates $T_P$. However, due to the equation duplication problem introduced in Section 1, this problem-equation templates formalization is a non-deterministic transduction that will have adverse effects on the performance of maximum likelihood estimation. There are two types of equation duplication: 1) order duplication such as ``$n1+n3+n2$'' and ``$n1+n2+n3$'', 2) bracket duplication such as ``$n_1+n_3-n_2$'' and ``$n1+(n_3-n_2)$''.

To normalize the order-duplicated templates, we define two normalization rules:

\begin{itemize}
\item{Rule 1: Two duplicated equation templates with unequal length should be normalized to the shorter one. For example, two equation templates ``$n_1+n_2+n_3+n_3 - n_3$'', ``$n_1+n_2+n_3$'' should be normalized to the latter one.}
\item{Rule 2: The number tokens in equation templates should be ordered as close as possible to their order in number mapping. For example, three equation templates ``$n_1+n_3+n_2$'', ``$n_1+n_2+n_3$'' and ``$n_3+n_1+n_2$'' should be normalized to ``$n_1+n_2+n_3$''.}
\end{itemize}

\begin{figure}[!htbp]
\centering
\includegraphics[width=1.5in]{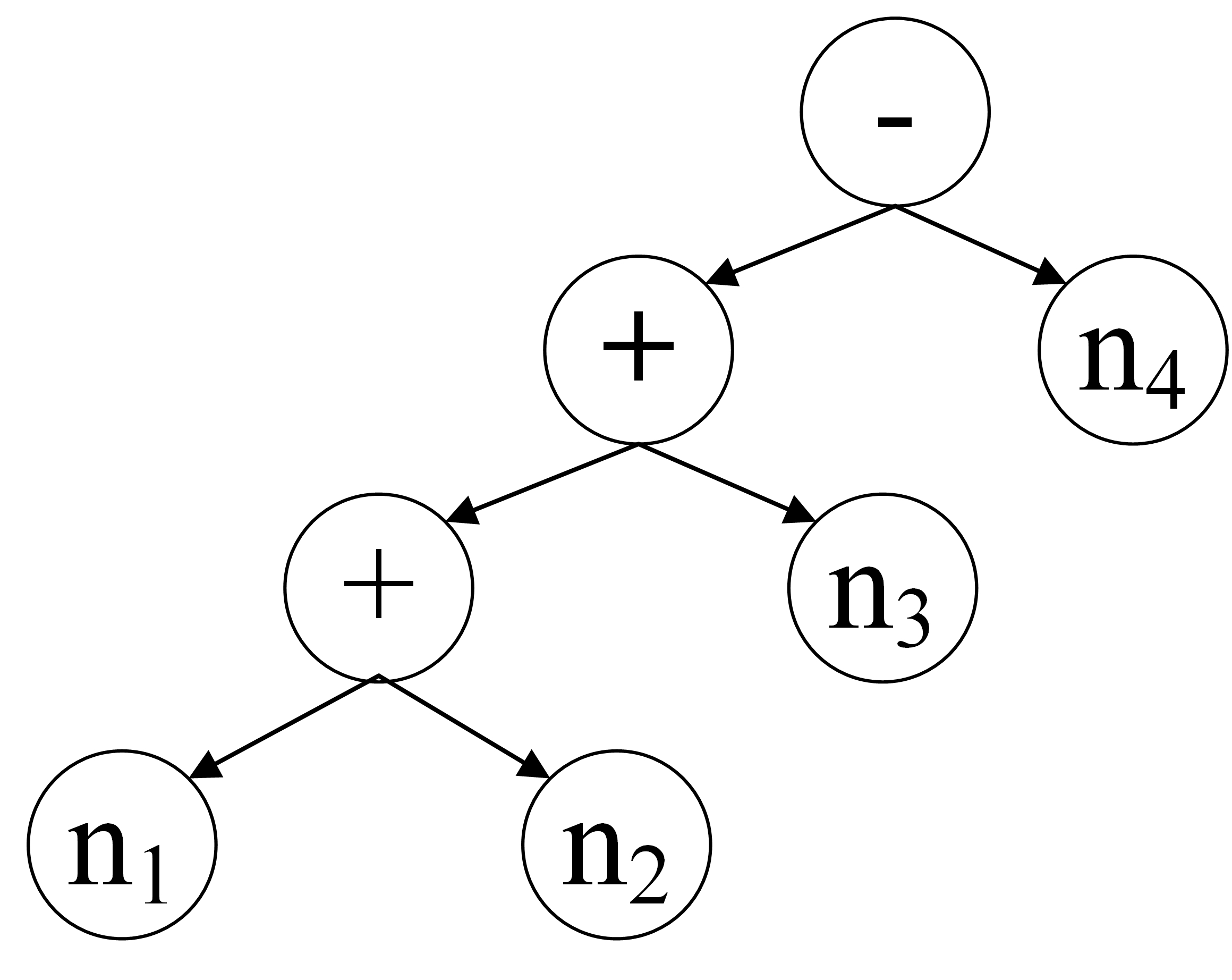}
\caption{A Unique Expression Tree}\label{tree_pic}
\end{figure}

To solve the bracket duplication problem, we further normalize the equation templates to an expression tree. Every inner node in the expression tree is an operator with two children, while each leaf node is expected to be a number token. An example of expressing equation template $n_1+n_2+n_3-n_4$ as the unique expression tree is shown in Figure \ref{tree_pic}.

\begin{figure}[!htbp]
\centering
\includegraphics[width=3in]{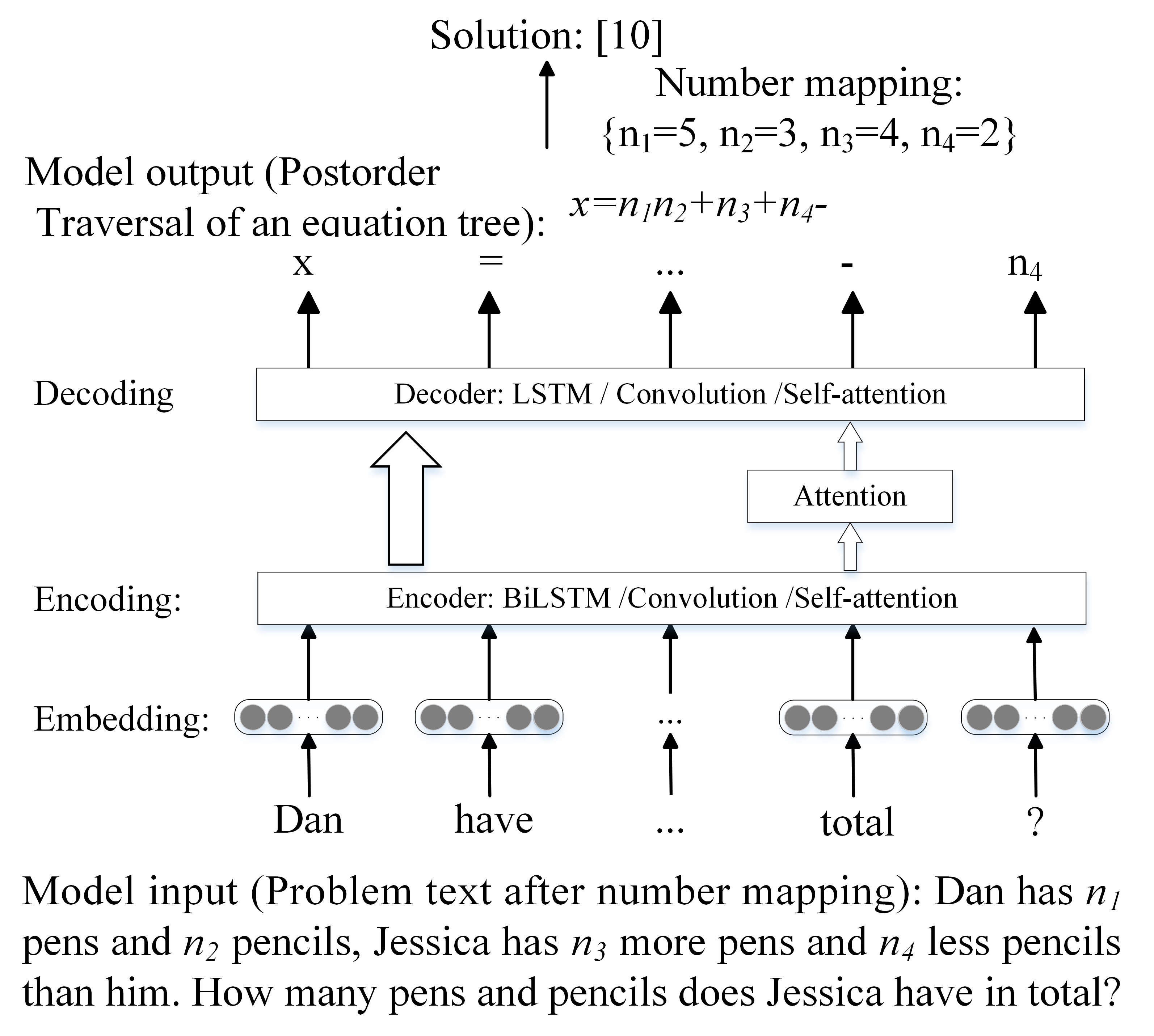}
\caption{Framework of \textsc{Seq2Seq} models}\label{seq2seq}
\end{figure}

After equation normalization, the \textsc{Seq2Seq} models can solve MWPs by taking problem text as source sequence and the postorder traversal of an unique expression tree as target sequence, as shown in Figure \ref{seq2seq}.

\section{Models}
In this section, we present three types of \textsc{Seq2Seq} models to solve MWPs: bidirectional Long Short Term Memory network (BiLSTM) \cite{wu2016google}, Convolutional \textsc{Seq2Seq} model \cite{gehring2017convolutional}, and Transformer \cite{vaswani2017attention}. To benefit the output accuracy with all three architectures, we propose to use a simple ensemble method.

\subsection{BiLSTM}

The BiLSTM model uses two LSTMs (forward and backward) to learn the representation of each token in the sequence based on both the past and the future context of the token. At each time step of decoding, the deocder uses a global attention mechanism to read those representations.

In more detail, we use two-layer Bi-LSTM cells with 256 hidden units as encoder, and two layers LSTM cells with 512 hidden units as decoder. In addition, we use Adam optimizer with learning rate $1e^{-3}$, $\beta_1 = 0.9$, and $\beta_2=0.99$. The epochs, minibatch size, and dropout rate are set to 100, 64, and 0.5, respectively.


\subsection{ConvS2S}
ConvS2S \cite{gehring2017convolutional} uses a convolutional architecture instead of RNNs. Both encoder and decoder share the same convolutional structure that uses $n$ kernels striding from one side to the other, and uses gate linear units as non-linearity activations over the output of convolution.

Our ConvS2S model adopts a four layers encoder and a three layers decoder, both using kernels of width 3 and hidden size 256. We adopt early stopping and learning rate annealing and set max-epochs equals to 100.

\subsection{Transformer}
\citet{vaswani2017attention} proposed the Transformer based on an attention mechanism without relying on any convolutional or recurrent architecture. Both encoder and decoder are composed of a stack of identical layers. Each layer contains two parts: a multi-head self-attention module and a position-wise fully-connected feed-forward network.

Our transformer is four layers deep, with $n_{head}=16$, $d_k = 12$, $d_v=32$, and $d_{model} = 512$, where $n_{head}$ is the number of heads of its self-attention, $d_k$ is the dimension of keys, $d_v$ is the dimension of values, and $d_{model}$ is the output dimension of each sub-layer. In addition, we use Adam optimizer with learning rate $1e^{-3}$, $\beta_1 = 0.9$, $\beta_2=0.99,$ and dropout rate of 0.3.

\subsection{Ensemble Model}
Through careful observation (detailed in Section 5.2), we find that each model has a speciality in solving problems. Therefore, we propose an ensemble model which selects the result according to models' generation probability:
$$p(\textbf{y}) = \prod_{t=1}^{T} p(y_t|y_{<t}, \textbf{x})$$
where $\textbf{y} = \{y_1,...,y_T\}$ is the target sequence, and $\textbf{x} = \{x_1,...,x_S\}$ is the source sequence. Finally, the output of the model with the highest generation probability is selected as the final output.




\section{Experiment}

In this section, we conduct experiments on dataset Math23K to examine the performance of different \textsc{Seq2Seq} models. Our main experimental result is to show a significant improvement over the baseline methods. We further conduct a case study to analyze why different \textsc{Seq2Seq} models can solve different kinds of MWPs.

 \textbf{Dataset:} Math23K\footnote{\url{https://ai.tencent.com/ailab/Deep_Neural_Solver_for_Math_Word_Problems.html}} collected by \citet{wang2017deep} contains 23,162 labeled MWPs. All these problems are linear algebra questions with only one unknown variable.

 \textbf{Baselines:} We compare our methods with two baselines: DNS and DNS-Hybrid. Both of them are proposed in \cite{wang2017deep}, with state-of-the-art performance on dataset Math23K. The DNS is a vanilla \textsc{Seq2Seq} model that adopts GRU \cite{chung2014empirical} as encoder and LSTM as decoder. The DNS-Hybrid is a hybrid model that combines DNS and a retrieval-based solver to achieve better performance.
%

\begin{table}[tp]
\centering
\begin{tabular}{c|c|c}

& Acc w/o EN (\%)& Acc w/ EN (\%) \\ \hline
DNS& 58.1 & 60.7\\ \hline
Bi-LSTM & 59.6 & 66.7\\
ConvS2S& 61.5 & 64.2\\
Transformer & 59.0 & 62.3 \\
Ensemble& 66.4& \textbf{68.4} \\
\end{tabular}
\caption{\label{exp-comparison} Model comparison. EN is short for equation normalization}
\end{table}

\begin{table}[tp]
\centering
\begin{tabular}{c|c|c|c}

& Bi-LSTM & Transformer & ConvS2S \\ \hline
w/o EN & 59.6 & 59.0 & 61.5 \\\hline
+ SE  & 63.1 & 59.9 & 62.2 \\
+ OE & 63.7 & 60.7 & 62.9 \\
+ EB & 65.3 & 61.2& 62.9 \\
\end{tabular}
\caption{\label{exp-ablation} The ablation study of three equation normalization methods. SE is the first Rule mentioned in Section 3. OE is the second rule mentioned in Section 3. EB means eliminating the brackets.}
\end{table}

\begin{table*}[htpb]
\centering
\begin{tabular}{|p{15cm}|}
\hline
 \textbf{Example 1:} Two biological groups have produced 690 ($n_1$) butterfly specimens in 15 ($n_2$) days. The first group produced 20 ($n_3$) each day. How many did the second group produced each day? \\ 

\textbf{Bi-LSTM:} $n_1 n_2 / n_3 -$; (correct)\quad \textbf{ConvS2S:} $n_2 n_3 + n_1 *$; (error) \quad \textbf{Transformer: }$n_2 n_1 n_3 n_3 +$; (error)\\ 
\hline
\hline
 \textbf{Example 2:} A plane, in a speed of 500 ($n_1$) km/h, costs 3 ($n_2$) hours traveling from city A to city B. It only costs 2 ($n_3$) hours for return. How much is the average speed of the plane during this round-trip? \\ 
 \textbf{Bi-LSTM: }$n_1 n_2 * n_3 *$; (error)\quad \textbf{ConvS2S: }$1 1 + 1 n_1 / 1 n_2 / + /$; (error)\quad \textbf{Transformer: }$ n_1 n_2 * n_3 * n_2 n_3 + /$; (correct)\\
\hline
\hline
\textbf{Example 3: } Stamp A is 2 ($n_1$) paise denomination, and stamp B is 7 ($n_2$) paise denomination. If we are asked to buy 10 ($n_3$) of each, how much more does it cost to buy stamps A than to buy stamps B.\\ 
 \textbf{Bi-LSTM: } $n_1 n_2 n_3 * - $; (error)\quad \textbf{ConvS2S: }$n_1 n_3 * n_2 n_3 * -$; (correct)\quad \textbf{Transformer: } $n_2 n_2 * n_2 n_3 * -$; (error)\\
\hline
\end{tabular}
\caption{\label{ex-examples} Three examples of solving MWP with \textsc{Seq2Seq} model. Note that the results are postorder traversal of expression trees, and the problems are translated to English for brevity.}
\end{table*}

\subsection{Results}
In experiments, we use the testing set in Math23K as the test set, and randomly split 1, 000 problems from the training set as validation set. Evaluation results are summarized in Table \ref{exp-comparison}. First, to examine the effectiveness of equation normalization, model performance with and without equation normalization are compared. Then the performance of DNS, DNS-Hybrid, Bi-LSTM, ConvS2S, Transformer, and Ensemble model are examined on the dataset.

Several observations can be made from the results. First, the equation normalization process significantly improves the performance of each model. The accuracy of different models gain increases from 2.7\% to 7.1\% after equation normalization. Second, Bi-LSTM, ConvS2S, Transformer can achieve much higher performance than DNS, which means that popular machine translation models are also efficient in automatic MWP solving. Third, by combining the \textsc{Seq2Seq} models, our ensemble model gains additional 1.7\% increase on accuracy.

In addition, we have further conducted three extra experiments to disentangle the benefits of three different EN techniques. Table \ref{exp-ablation} gives the details of the ablation study of the three \textsc{Seq2seq} models. Taking Bi-LSTM as an example, accuracies of rule 1 (SE), rule 2 (OE) and eliminating brackets (EB) are 63.1\%, 63.7\% and 65.3\%, respectively. Obviously, the performance of \textsc{Seq2esq} models benefits from the equation normalization technologies.

\subsection{Case Study}

Further, we conduct a case analysis on the capability of different \textsc{Seq2seq} models and provide three examples in Table \ref{ex-examples}. Our analysis is summarized as follows: 1) Transformer occasionally generates mathematically incorrect templates, while Bi-LSTM and ConvS2S almost do not, as shown in Example 1. This is probably because the size of training data is still not enough to train the multi-head self-attention structures; 2) In Example 2, the Transformer is adapted to solve problems that require complex inference. It is mainly because different heads in a self-attention structure can model various types of relationships between number tokens; 3) The multi-layer convolutional block structure in ConvS2S can properly process the context information of number tokens. In Example 3, it is the only one that captures the relationship between stamp A and stamp B.

\section{Conclusion}
In this paper, we first propose an equation normalization method that normalizes duplicated equation templates to an expression tree. We test different \textsc{seq2seq} models on MWP solving and propose an ensemble model to achieve higher performance. Experimental results demonstrate that the proposed equation normalization method and the ensemble model can significantly improve the state-of-the-art methods.

\section*{Acknowledgement}
This work is supported in part by the National Nature Science Foundation of China under grants No. 61602087, and the Fundamental Research Funds for the Central Universities under grants No. ZYGX2016J080.
\bibliography{math_word_problem_5_20}
\bibliographystyle{acl_natbib_nourl}

\end{document}